\pdfoutput=1

\documentclass[11pt]{article}

\usepackage[]{acl}
\usepackage{multirow}
\usepackage{times}
\usepackage{latexsym}
\usepackage{graphicx}
\usepackage{amsmath}
\usepackage{booktabs}
\usepackage{CJKutf8}
\usepackage{arydshln}
\usepackage{tikz}
\usepackage{makecell}
\usepackage{color}
\usepackage{tabularx}
\usepackage{bm}
\graphicspath{ {images/} }
\usepackage{pgfplots, pgfplotstable}
\usepgfplotslibrary{groupplots}
\usepackage{tikz}
\usepackage[T1]{fontenc}

\usepackage[utf8]{inputenc}

\usepackage{microtype}

\title{Joint Alignment of Multi-Task Feature and Label Spaces\\ for Emotion Cause Pair Extraction}

\author{Shunjie Chen\textsuperscript{$1$}, Xiaochuan Shi\textsuperscript{$1$}, Jingye Li$^1$, Shengqiong Wu$^2$, Hao Fei$^2$, Fei Li\textsuperscript{$1\dagger$}, Donghong Ji$^1$ \\
$^1$Key Laboratory of Aerospace Information Security and Trusted Computing, Ministry of \\Education, School of Cyber Science and Engineering, Wuhan University, Wuhan, China \\ 
$^2$School of Computing, National University of Singapore, Singapore\\
\texttt{\{shunjiechen, shixiaochuan, theodorelee, lifei\_csnlp, dhji\}@whu.edu.cn}\\
\texttt{swu@u.nus.edu, haofei37@nus.edu.sg}
}

\begin{document}
\begin{CJK}{UTF8}{gbsn}
\maketitle
\renewcommand{\thefootnote}{\fnsymbol{footnote}}
\footnotetext[2]{Corresponding author}
\renewcommand{\thefootnote}{\arabic{footnote}}
\begin{abstract}
Emotion cause pair extraction (ECPE), as one of the derived subtasks of emotion cause analysis (ECA), shares rich inter-related features with emotion extraction (EE) and cause extraction (CE).
Therefore EE and CE are frequently utilized as auxiliary tasks for better feature learning, modeled via multi-task learning (MTL) framework by prior works to achieve state-of-the-art (SoTA) ECPE results.
However, existing MTL-based methods either fail to simultaneously model the specific features and the interactive feature in between,
or suffer from the inconsistency of label prediction.
In this work, we consider addressing the above challenges for improving ECPE by performing two alignment mechanisms with a novel A$^2$Net model.
We first propose a feature-task alignment to explicitly model the specific emotion-\&cause-specific features and the shared interactive feature.
Besides, an inter-task alignment is implemented, in which the label distance between the ECPE and the combinations of EE\&CE are learned to be narrowed for better label consistency.
Evaluations of benchmarks show that our methods outperform current best-performing systems on all ECA subtasks.
Further analysis proves the importance of our proposed alignment mechanisms for the task.\footnote{Our code is available at  \url{https://github.com/csj199813/A2Net_ECPE}}

\end{abstract}

\section{Introduction}

\begin{figure}[!t]
    \centering
    \includegraphics[width=0.9\columnwidth]{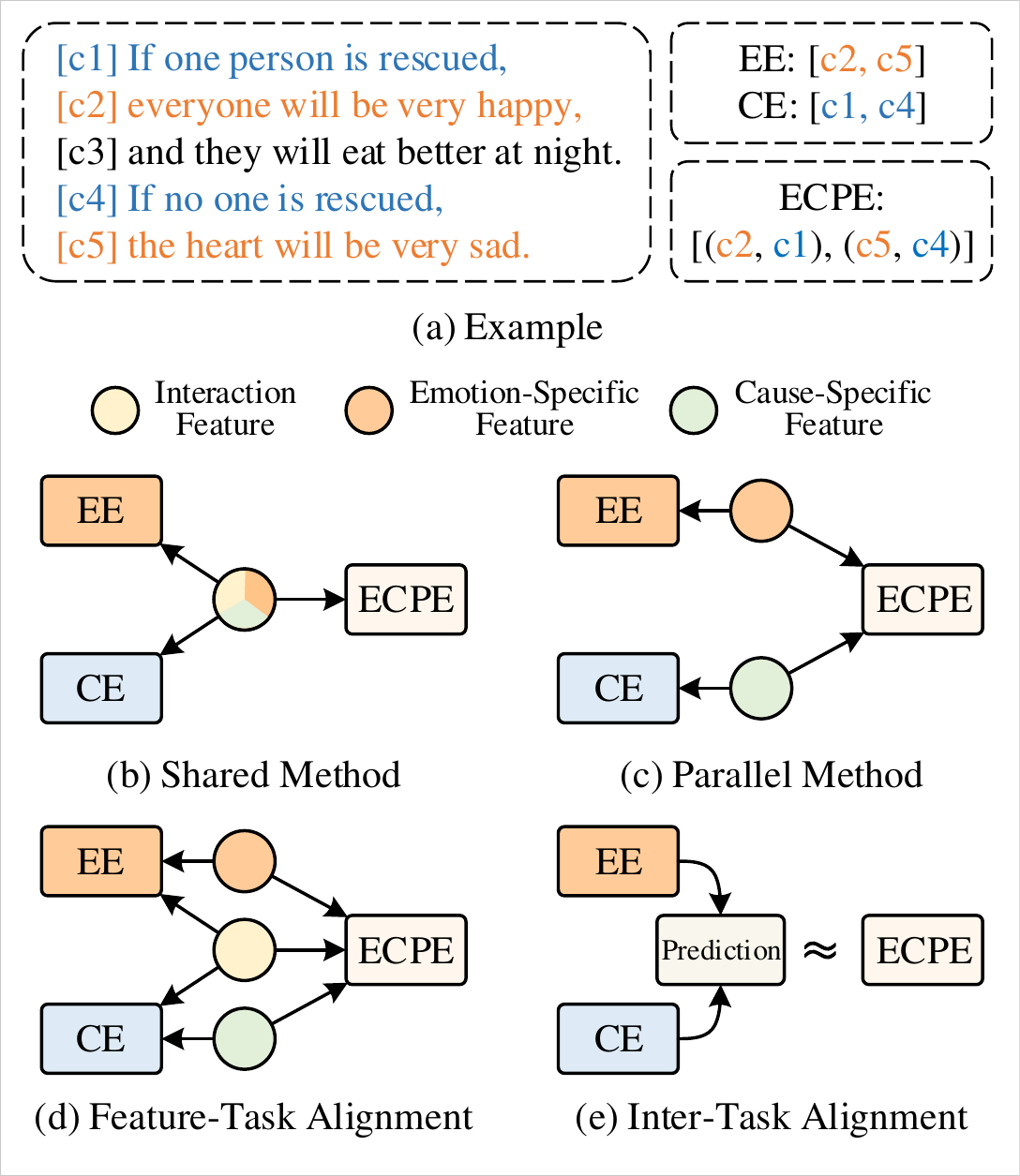}
    \caption{(a) illustrates three subtasks of ECA. 
    (b) and (c) depicts the shared and parallel features encoding method, respectively.
    In (d) and (e) we show our proposed feature-task alignment mechanism and inter-task alignment mechanism, respectively.}
    \label{fig:introduction}
\end{figure}

Emotion cause analysis (ECA), detecting potential causes for certain emotion expressions in a document, has been a hot research topic in natural language processing (NLP) community \cite{lee2010text, gui-etal-2016-event,fan-etal-2019-knowledge}.
ECA has derived three associated tasks: EE, CE and ECPE.
As illustrated in Figure \ref{fig:introduction}(a), EE and CE detects the emotion and cause clauses respectively, while ECPE identifies both the emotion and cause clauses as well as their semantic relation.
By jointly modeling the clauses detection and the relational pairing, ECPE effectively relieves the noise introduction in the pipeline process, and thus receives most research attention recently \cite{ding2020ecpe,wei2020effective,Wu0LZLTJ22}.

As there are close correlations among EE, CE and ECPE, existing ECPE works extensively treat EE and CE as auxiliary tasks for additional feature supports, and mostly adopt the multi-task learning framework to explicitly model the inter-dependency in between, thus achieving current SoTA performances \cite{wei2020effective,ding2020ecpe,fan2021order,bao2022multi}. 
From the view of feature encoding, there are two major categories of MTL-based ECPE methods: \emph{shared feature encoding} method and \emph{parallel feature encoding} method.
As shown in Figure \ref{fig:introduction}(b), shared methods only learn mixed features via one encoder without distinguishing specific features for individual subtasks \cite{wei2020effective,yuan-etal-2020-emotion}.
In contrast, parallel methods \cite{ding2020ecpe,ding2020end,fan2021order} use two encoders to learn emotion- and cause-specific features separate (cf. Figure \ref{fig:introduction}(c)), where unfortunately, the interaction among these tasks are overlooked.
We argue that both the private emotion-specific and cause-specific features and the shared interactive feature are important to the final performance, which should be explicitly modeled in the MTL framework.
To this end, we in this work propose a \emph{feature-task alignment} (FTA) scheme of MTL for ECPE (cf. Figure \ref{fig:introduction}(d)), in which we explicitly split three parts of features, and align them to EE, CE and ECPE respectively.

Meanwhile, aligning the label space in a MTL framework is crucial to overall ECPE performance, because intuitively all modules in the MTL process should reach a consensus. 
For example as in Figure \ref{fig:introduction}(a), once ``c1'' is recognized as a cause clause by ECPE module, it should not be further predicted as a non-cause clause by CE module.
We notice that such label consistency is not guaranteed in existing MTL-based ECPE methods, which could inevitably hurt the prediction.
Therefore we further introduce a \emph{inter-task alignment} (ITA) mechanism (cf. \ref{fig:introduction}(e)) that learns to pull closer the label distance between the ECPE and the combinations of EE\&CE, ensuring label consistency.

We implement the above ideas of feature-task \underline{a}lignment and inter-task \underline{a}lignment by developing a novel neural network, namely A$^2$Net, as shown in Figure \ref{fig:ecpe}.
First, we employ the BERT \cite{devlin-etal-2019-bert}
as document encoder for producing clauses representations.
We then leverage the partition filter network (PFN) \cite{yan2021partition} to implement the feature-task alignment, generating emotion-specific features, cause-specific features and interaction features, respectively.
Afterwards, we apply emotion-specific and interaction features for EE, cause-specific and interaction features for CE, and all features for ECPE.
Finally, we reach the goal of inter-task alignment by minimizing the bidirectional KL-divergence between the output distributions of ECPE and EE$\times$CE, thus maintaining the consistency of label spaces among all tasks.

Our A$^2$Net framework is evaluated on the ECA benchmark \cite{xia-ding-2019-emotion}, where our system achieves new SoTA results on EE, CE and ECPE.
Further analyses demonstrate that our method learns better consistency in the predictions of all subtasks than existing baselines.
Overall, this work contributes to three major aspects:

\begin{itemize}
\setlength{\itemsep}{2pt}
\setlength{\parsep}{2pt}
\setlength{\parskip}{2pt}
    \item We present an innovative multi-task learning based ECPE framework, where we further propose a feature-task alignment mechanism that can make better use of the shared features from EE and CE sources.

    \item We also introduce an inter-task alignment mechanism to reduce the inconsistency between the prediction results of ECPE and the EE\&CE, significantly enhancing the performance as well as the robustness of the system.

    \item Our system empirically achieves new SoTA performances of the EE, CE and ECPE tasks on the benchmark.
    
\end{itemize}

\section{Related Work}

In NLP area, the analysis on sentiment and opinion is a long-standing research topic \cite{2012Liu,LiFJ20,Wu0RJL21,0001LJL22}, including detecting of the sentiment polarities \cite{TangQL16,Feiijcai22UABSA,shi-etal-2022-effective} and the emotion categories \cite{lee2010text,neviarouskaya2013extracting}.
One of the recent trend on the emotion detection has been upgraded to the emotion cause analysis (ECA).
Centered on the topic of ECA, there are several subordinated tasks according to the extracting elements of emotion and cause, such as emotion cause extraction (ECE) \cite{lee2010text,neviarouskaya2013extracting,gui-etal-2016-event,li2018co} and ECPE \cite{xia-ding-2019-emotion, wei2020effective,bao2022multi}.

\citet{lee2010text} pioneer the ECE task, in which the task is formulated as a word-level cause labeling problem. 
Following this work, initial research constructs rule-based methods \cite{neviarouskaya2013extracting,gao2015emotion} and machine learning methods \cite{ghazi2015detecting, song2015detecting} on their own corpus.
Deep learning based methods greatly facilitate the line of this research \cite{0001LLJ21,Feiijcai22DiaSRL,Feiijcai22DiaRE,0001ZLJ21,HashtagRec}.
Recently, \citet{gui-etal-2016-event} release a public corpus and re-formalize ECE as a clause-level classification problem, where the goal is to detect cause clauses for a given emotion in the text.
The framework has received much attention in recent years and the corpus has become a benchmark ECA dataset \cite{gui-etal-2017-question,li2018co,fan-etal-2019-knowledge,ding2019independent, hu2021fss,yan-etal-2021-position,hu-etal-2021-bidirectional-hierarchical}.

However, as \citet{xia-ding-2019-emotion} points out, the ECE task is limited to the task definition, i.e., emotion needs to be manually marked in advance.
They thus introduce the ECPE task that simultaneously extract both the emotion and cause clauses as well as determining their relations, which has a better utility in real-world applications \cite{0001WRZ22,fei-etal-2020-cross}.
Thereafter, a line of subsequent research efforts are paid to ECPE within the last years \cite{ding2020end,cheng2020symmetric,fan2021order,bao2022multi}.

Recent ECPE methods mostly employ the multi-task learning for simultaneously modeling the EE and CE as auxiliary tasks for making use of the shared features, and thus realize SoTA ECPE performances \cite{bao2022multi}.
Existing MTL-based ECPE works can largely be divided into two categories: parallel encoding and shared encoding methods. 
\emph{Parallel methods} mostly learn the emotion/cause feature representations in mutually independent ways \cite{cheng2020symmetric,ding2020ecpe,fan2021order,bao2022multi}.
\citet{ding2020end,chen2022recurrent} uses auxiliary task prediction to aid the interactions between emotion and cause features.
However, the prediction values are limited to only a two-dimensional vector, leading to insufficient interaction between emotion and cause features.

\emph{Shared methods} learn the mixed features by only one encoder without distinguishing between features of different tasks \cite{wei2020effective, yuan-etal-2020-emotion,NMCL,MMGCN}.
In this work, we use the PFN \cite{yan2021partition} to generate emotion-specific features, cause-specific features and interaction features and implement the feature-task alignment.
Moreover, we use an inter-task alignment module to reduce the gap between EE, CE and ECPE, maintaining the consistency of the label space.

\section{Methodology}

\paragraph{Task Formulation} 
Given a document consisting of $N$ clauses $\mathcal{D} = \{c_1, c_2, \dots, c_N\}$, and each $c_i$ denotes a subsection of a sentence separated by a comma.
The goals of EE and CE task are extracting emotion clauses $c_i^{e} \in \mathcal{D}$ and cause clauses $c_j^{c} \in \mathcal{D}$ in the document, respectively, while ECPE task identifies the emotion-cause clause pair ($c_i^{e}$, $c_j^{c}$) that has causal relationship between emotion and cause clauses.

\begin{figure*}[htp]
    \centering
    \includegraphics[width=1.0\textwidth]{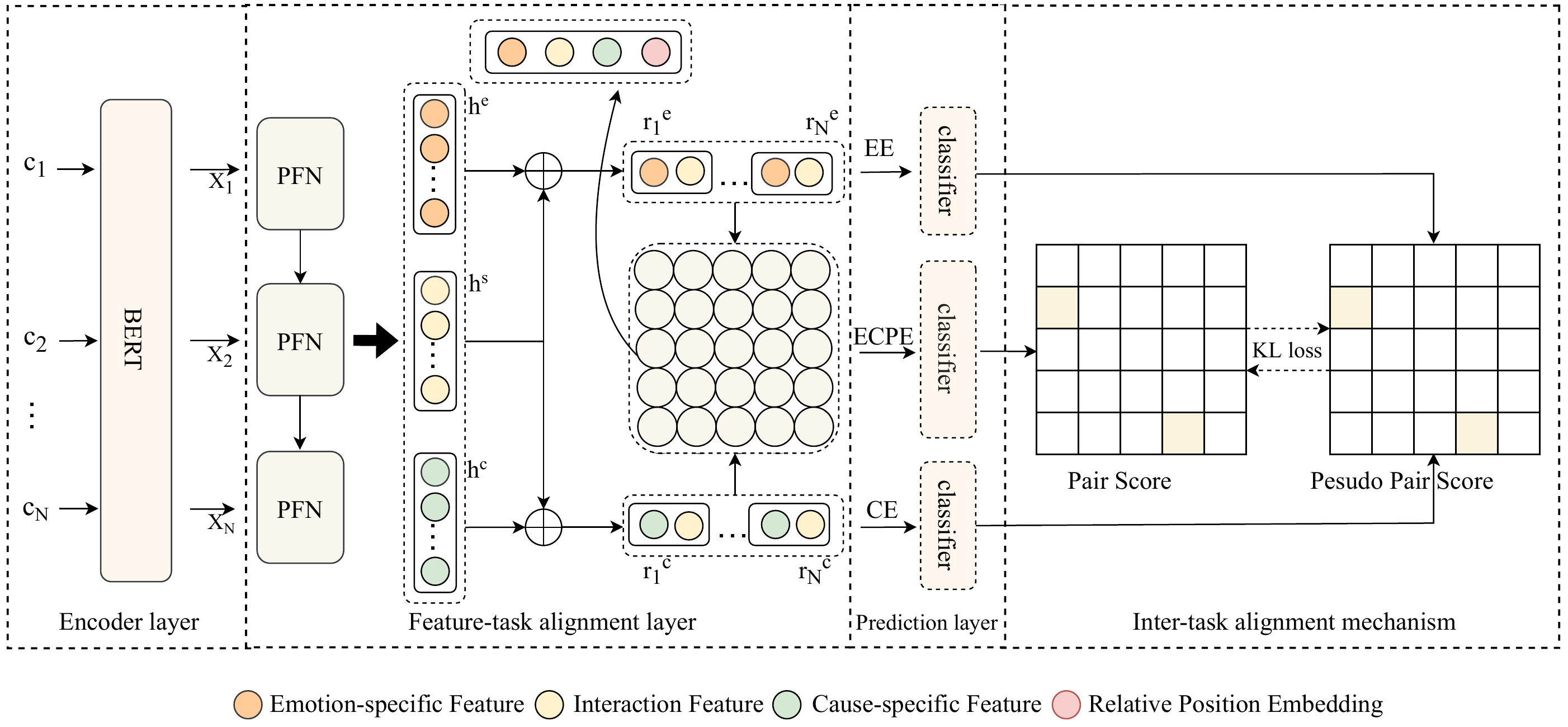}
    \caption{Overview of our A$^2$Net model.}
    \label{fig:ecpe}
\end{figure*}

As illustrated in Figure \ref{fig:ecpe}, the overall architecture of our A$^2$Net consists of four tiers, including the encoder layer, feature-task alignment layer, prediction layer and inter-task alignment mechanism.
First, following the previous work and we use BERT \cite{devlin-etal-2019-bert} as encoder to yield contextualized clause representations from input documents.
Then, to explicitly model task-specific features and task-shared features.
We leverage a PFN \cite{yan2021partition} to capture emotion- and cause-specific features and the interaction between them.
Afterward, a prediction layer is used to align three combinations of three kinds of features from PFN with three tasks, and predict the emotion clauses, cause clauses and emotion-cause pair for EE, CE and ECPE, respectively.
Finally, considering that there should be a consensus among all tasks. 
We propose an inter-task alignment mechanism to enhance the consistency between ECPE and EE$\times$CE.

\subsection{Encoder Layer}

Following \cite{wei2020effective}, we also leverage pre-trained BERT language model \cite{devlin-etal-2019-bert} as the underlying encoder to yield contextualized clause representations.
Concretely, we insert a \verb|[CLS]| token at the beginning of each clause and append a \verb|[SEP]| token to the end, i.e., $c_i = \{[CLS], w_{i,1}, w_{i,2},...,w_{i,M}, [SEP]\}$.
Then we concatenate them together as the input of BERT to generate contextualized token representations,
in which we take the representation of \verb|[CLS]| token in each clause $c_i$ as its clause representation.
After that, we obtain all the clause representations $\bm{X} = \{\bm{x}_1, \bm{x}_2, \dots, \bm{x}_N\}$.

\subsection{Feature-task Alignment Layer}

We adopt partition filter network (PFN) \cite{yan2021partition} to capture emotion- and cause-specific features and the interaction between them because of its powerful ability to extract task-specific features and interaction features.
PFN is similar to the LSTM structure and has two task-related gates: the emotion gate and the cause gate. 
The gates filter features according to their contribution to each task with emotion and cause gates.
In each time step, the encoder divides clause representation into three feature partitions: emotion partition, cause partition, and interaction partition, where interaction partition represents information useful to all tasks. 

Specifically, 
at the \textit{i} time step, we first generate two task-related gates:
\begin{equation}\label{gate}
\begin{aligned}
&\bm{g}^{e}_i = \text{Cummax}(\text{Linear}([\bm{x}_i;\bm{h}_{i-1}])) \,, \\
&\bm{g}^{c}_i = 1 - \text{Cummax}(\text{Linear}([\bm{x}_i;\bm{h}_{i-1}])) \,,
\end{aligned}
\end{equation}
where
$\text{Cummax}(\cdot)=\text{Cumsum}(\text{Softmax}(\cdot))$, performs as a binary gate, and $\text{Linear}(\cdot)$ denotes linear transformation, and $\bm{h}_{i-1}$ is the hidden state of ${i}$-1-th clause.
Each gate will divide clause representations into two segments: task-related and task-unrelated, according to their usefulness to the specific task.
With the joint efforts of the two gates, the clause representation can be divided into three partitions: emotion partition $\bm{p}^e_{i}$, cause partition $\bm{p}^c_{i}$ and interaction partition $\bm{p}^s_{i}$.
We use task-related gates ($\bm{g}^{c}_i$ and $\bm{g}^{e}_i$) to calculate forgetting gates:

\begin{equation}
\label{partition}
\begin{aligned}
\bm{f}^s_{i} &= \bm{g}^{e}_i \circ \bm{g}^{c}_i \,, \\
\bm{f}^e_{i} &= \bm{g}^{e}_i - \bm{f}^s_{i} \,, \\
\bm{f}^c_{i} &= \bm{g}^{c}_i - \bm{f}^s_{i} \,,
\end{aligned}
\end{equation}
where $\circ$ denotes element-wise multiplication, $\bm{f}^e_{i}$, $\bm{f}^c_{i}$ and $\bm{f}^s_{i}$ are emotion, cause and interaction forgetting gates, respectively.
Similarly, we also can obtain input gates $\bm{o}^e_{i}$, $\bm{o}^c_{i}$ and $\bm{o}^s_{i}$ via Equation \ref{gate} and \ref{partition}.

After that, we use forgetting and input gates to control the flow of input and history information:
\begin{equation}
\begin{aligned}
&\widetilde{\bm{c}}_{i} = \text{tanh}(\text{Linear}([\bm{x}_i;\bm{h}_{i-1}])) \,, \\
&\bm{p}^s_{i} = \bm{f}^s_{i} \circ \bm{c}_{i-1} + \bm{o}^s_{i} \circ \widetilde{\bm{c}}_{i} \,, \\
&\bm{p}^e_{i} = \bm{f}^e_{i} \circ \bm{c}_{i-1} + \bm{o}^e_{i} \circ \widetilde{\bm{c}}_{i} \,, \\
&\bm{p}^c_{i} = \bm{f}^c_{i} \circ \bm{c}_{i-1} + \bm{o}^c_{i} \circ \widetilde{\bm{c}}_{i} \,,
\end{aligned}
\end{equation}
where $\widetilde{\bm{c}}_{i}$ denotes the current input information, and $\bm{c}_{i-1}$ denotes the history information.

Next, we can obtain three feature representations: emotion feature $\bm{h}^e_{i}$, cause feature $\bm{h}^c_{i}$ and inter-task interaction feature $\bm{h}^s_{i}$ from the partition:
\begin{equation}
\begin{aligned}
\bm{h}^s_{i} &= \text{tanh}(\bm{p}^s_{i}) \,, \\
\bm{h}^e_{i} &= \text{tanh}(\bm{p}^e_{i}) \,, \\
\bm{h}^c_{i} &= \text{tanh}(\bm{p}^c_{i}) \,.
\end{aligned}
\end{equation}

We further use the information in all three partitions to construct cell state $\bm{c}_{i}$, and hidden state $\bm{h}_i$ for the next time step:
\begin{equation}
\begin{aligned}
\bm{c}_{i} &= \text{Linear}([\bm{p}^e_{i};\bm{p}^s_{i};\bm{p}^c_{i}]) \,, \\
\bm{h}_i &= \text{tanh}(\bm{c}_{i}) \,.
\end{aligned}
\end{equation}

After PFN, we can obtain the emotion-specific features $\bm{h}^e_{i}$, cause-specific features $\bm{h}^c_{i}$ and the interaction features $\bm{h}^s_{i}$.
First, we align features with the EE and CE tasks, in which we concatenate the features of emotion and cause with the interaction features separately and gain the emotion representations $\bm{r}^{e}_i = [\bm{h}^e_{i}; \bm{h}^s_{i}]$ and the cause representations $\bm{r}^{c}_i = [\bm{h}^c_{i}; \bm{h}^s_{i}]$.
Moreover, we consider aligning features with ECPE task, so we add task-specific features and interaction features to aggregate all the information about ECPE task:
\begin{equation}
\begin{aligned}
\bm{h}^{e'}_{i} &= \bm{h}^e_{i} + \bm{h}^s_{i} \,, \\
\bm{h}^{c'}_{j} &= \bm{h}^c_{j} + \bm{h}^s_{j}  \,, \\
\bm{r}_{ij} &= [\bm{h}^{e'}_{i};\bm{h}^{c'}_{j};\bm{e}_{ij}] \,, 
\end{aligned}
\end{equation}
where $\bm{e}_{ij}$ denotes the relative position embedding following \citet{wei2020effective}. $\bm{r}_{ij}$ is the final emotion-cause pair representation.

\subsection{Prediction Layer}
\paragraph{Extracting Emotion/Cause}

We feed $\bm{r}^{e}_i$ and $\bm{r}^{c}_i$ into two feedforward network (FFN) to obtain emotion prediction $\hat{y}^{e}_{i}$ and cause prediction $\hat{y}^{c}_{i}$ for the $i$-th clause:
\begin{equation}\label{aux prediction}
\begin{aligned}
\hat{y}^{e}_{i} &= \sigma(\text{FFN}(\bm{r}^{e}_i)) \,, \\
\hat{y}^{c}_{i} &= \sigma(\text{FFN}(\bm{r}^{c}_i)) \,,
\end{aligned}
\end{equation}
where $\sigma(\cdot)$ means the sigmoid function.

The auxiliary task loss for emotion prediction and cause prediction can be formulated as:
\begin{equation}
\mathcal{L}_{aux} = -\sum^N_{i=1}({y}^{e}_{i}\log(\hat{y}^{e}_{i})+{y}^{c}_{i}\log(\hat{y}^{c}_{i})) \,,
\end{equation}
where ${y}^{e}_{i}$ and ${y}^{c}_{i}$ are emotion and cause ground truth labels of clause $c_i$, respectively.

\paragraph{Extracting Emotion Cause Pair}

We employ a FFN with a sigmoid function to obtain the emotion-cause pair score $\hat{y}^{p}_{ij}$:
\begin{equation}
    \hat{y}^{p}_{ij} = \sigma(\text{FFN}(\bm{r}_{ij}))  \,.
\end{equation}

The loss function of emotion-cause pair extraction can be formalized as:
\begin{equation}
\mathcal{L}_{pair} = -\sum^N_{i=1}\sum^N_{j=1}{y}^{p}_{ij}\log(\hat{y}^{p}_{ij}) \,, 
\end{equation}
where ${y}^{p}_{ij}$ is the ground truth label of the clause pair $(c_i, c_j)$.

\subsection{Inter-task Alignment Mechanism}

\begin{figure}[!tp]
    \centering
    \includegraphics[width=7.5cm]{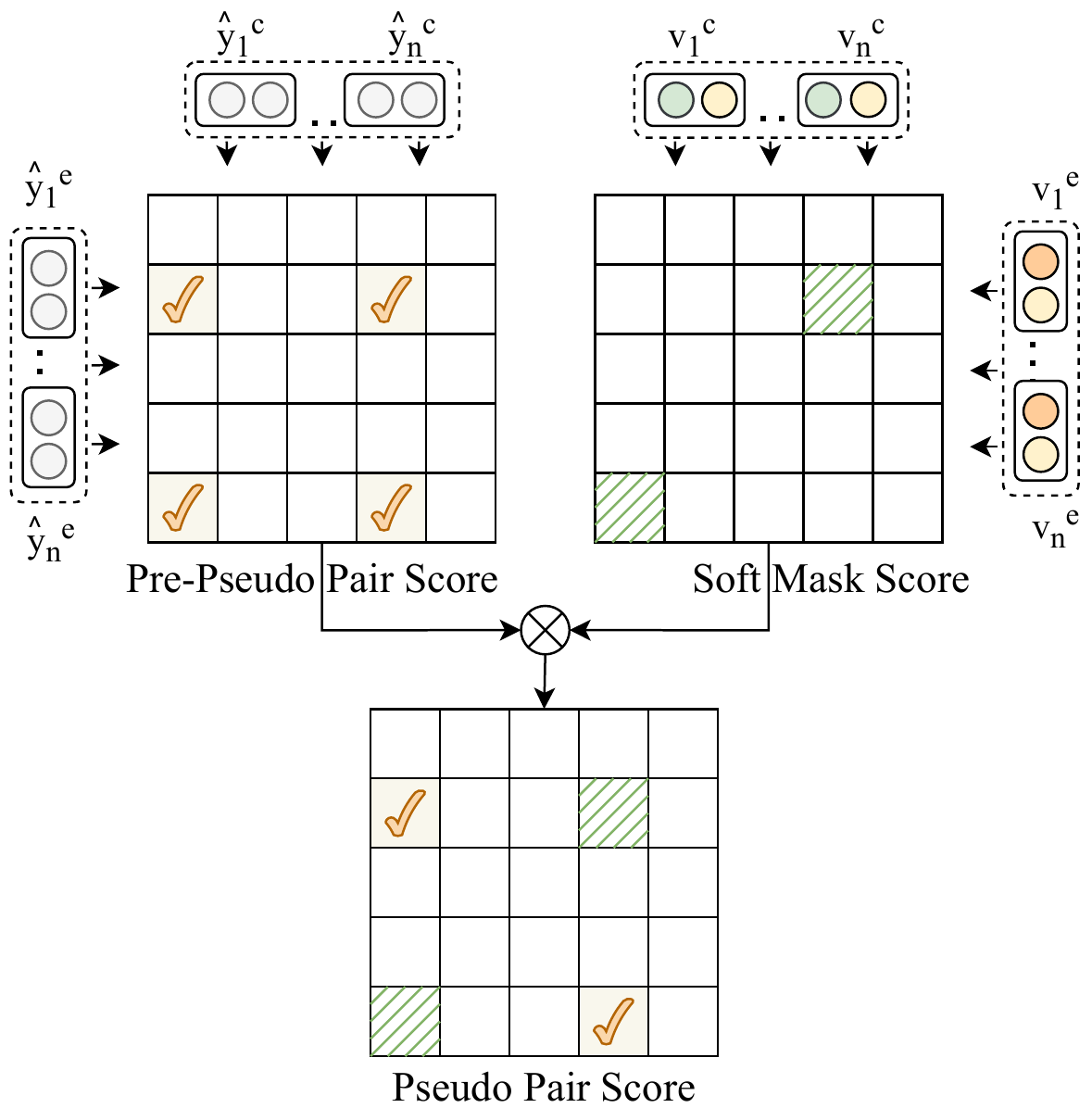}
    \caption{The generation of pseudo pair score, where √ denotes candidate emotion-cause pairs, green grids represent the masked pair.}
    \label{fig:alignment}
\end{figure}

As we argued earlier, the predictions of EE and CE could be inconsistent with ECPE, i.e., the emotion in the emotion-cause pair predicted by ECPE could not be detected by EE, which hinders the task for further improvements.
To address this issue, we propose an inter-task alignment (ITA) mechanism to constrain the predicted scores between ECPE and auxiliary tasks during the training period.
First, we leverage the emotion score $\hat{y}^{e}_{i}$ and the cause score $\hat{y}^{c}_{j}$ to get the pre-pseudo pair score $\sqrt{\hat{y}^{e}_{i}\hat{y}^{c}_{j}}$.
Note that there could not exist a causal relationship in pairs matched from predictions of EE and CE.
Thus, as shown in Figure \ref{fig:alignment}, we calculate the pseudo emotion-cause pair score $\tilde{y}_{ij}^{p}$ as follows:
\begin{equation}
 \tilde{y}_{ij}^{p} = \alpha_{ij}\sqrt{\hat{y}^{e}_{i}\hat{y}^{c}_{j}} \,, 
\end{equation}
where $\alpha_{ij}$ ($0 \leq \alpha_{ij} \leq 1$) is a soft mask score for the pseudo pair ($\hat{y}^{e}_{i}$, $\hat{y}^{c}_{j}$), which can reduce the score of fake emotion-cause pairs in pre-pseudo pairs score.
$\alpha_{ij}$ is computed by:
\begin{equation}
\begin{aligned}
    \bm{t}_{ij} &=  \frac{(\bm{v}^{e}_{i})^\top \bm{v}^{c}_{j}}{\sqrt{d}} \,, \\
    \alpha_{ij} &= \frac{\exp(\bm{t}_{ij})}{\sum^{N}_{j} \exp(\bm{t}_{ij}) } \,,
\end{aligned}
\end{equation}
where $\bm{v}^{e}_{i}$ and $\bm{v}^{c}_{j}$ are obtained from $\bm{r}^{e}_i = [\bm{h}^e_{i}; \bm{h}^s_{i}]$ and $\bm{r}^{c}_j = [\bm{h}^c_{j}; \bm{h}^s_{j}]$ with FFNs, respectively. 
The $d$ denotes the dimension of $\bm{v}^{e}_{i}$ and $\bm{v}^{c}_{j}$.

Then we reduce the gap between the pseudo emotion-cause pair score $\tilde{y}_{ij}^{p}$ from EE and CE and the true emotion-cause pair score $\hat{y}^{p}_{ij}$ from ECPE using Kullback Leibler (KL) Divergence:
\begin{equation}
\begin{aligned}
\mathcal{L}_{KL} &= \frac{1}{2}\sum^N_{i=1}\sum^N_{j=1}(\text{KL}(\tilde{y}_{ij}^{p}||\hat{y}^{p}_{ij}) + \text{KL}(\hat{y}^{p}_{ij}||\tilde{y}_{ij}^{p})) \, \\
&= \frac{1}{2}(\sum^N_{i=1}\sum^N_{j=1}(\tilde{y}_{ij}^{p}\log(\frac{\tilde{y}_{ij}^{p}}{\hat{y}^{p}_{ij}}) + \hat{y}^{p}_{ij}\log(\frac{\hat{y}^{p}_{ij}}{\tilde{y}_{ij}^{p}})) \,.
\label{KL}
\end{aligned}
\end{equation}

\paragraph{Optimization}

The final loss of our model is a weighted sum of $\mathcal{L}_{pair}$, $\mathcal{L}_{aux}$ and $\mathcal{L}_{KL}$ :
\begin{align}
\mathcal{L} &= \mathcal{L}_{pair} + \lambda_1\mathcal{L}_{aux} + \lambda_2\mathcal{L}_{KL} \,, 
\end{align}
where $\lambda_1$ and $\lambda_2$ are hyperparameters.

\begin{table*}[!t]
\centering
\begin{tabular}{lccccccccccccc}
\toprule
\multirow{2}*{\textbf{Approach}}& \multicolumn{3}{c}{\textbf{ECPE}}& \multicolumn{3}{c}{\textbf{EE}} &\multicolumn{3}{c}{\textbf{CE}}\\
\cmidrule(r){2-4} \cmidrule(r){5-7} \cmidrule(r){8-10}
\multirow{2}*{} & P & R & F1 & P & R & F1 & P & R & F1 \\
\midrule
ANTS & 72.43 & 63.66 & 67.76 & 81.96 & 73.29 & 77.39 & 74.90 & 66.02 & 70.18 \\
TransECPE & 73.74 & 63.07 & 67.99 & 87.16 & 82.44 & 84.74 & 75.62 & 64.71 &	69.74 \\
ECPE-2D & 72.92 & 65.44 & 68.89 & 86.27 & \textbf{92.21} & 89.10 & 73.36 & 69.34 & 71.23 \\
PairGCN & 76.92 & 67.91 & 72.02 & 88.57	& 79.58 & 83.75 & \underline{79.07} & 69.28 & 73.75 \\
RANKCP & 71.19 & \underline{76.30} & 73.60 & \underline{91.23} & 89.99 & \underline{90.57} & 74.61 & 77.88 & 76.15 \\
ECPE-MLL & \underline{77.00} & 72.35 & 74.52 & 86.08 & \underline{91.91} & 88.86 & 73.82 & \underline{79.12} & 76.30 \\
MGSAG & \textbf{77.43} & 73.21 & \underline{75.21} & \textbf{92.08} & 82.11 & 87.17 & \textbf{79.79} & 74.68 & \underline{77.12} \\
\midrule
A$^2$Net(ours) & 75.03 & \textbf{77.80} & \textbf{76.34} & 90.67 & 90.98 & \textbf{90.80} & 77.62 & \textbf{79.20} & \textbf{78.35} \\
\bottomrule
\end{tabular}
\caption{Comparisons with baselines on Chinese benchmark ECPE corpus. For a fair comparison, they all use BERT as the encoder. The best performance is in \textbf{bold} and the second best performance is \underline{underlined}. }\label{tab:main reuslt}
\end{table*}

\section{Experiments Settings}

\subsection{Dataset and Metrics}

We conducted experiments on the Chinese benchmark dataset provided by \citet{xia-ding-2019-emotion} to evaluate the effectiveness of our proposed model A$^2$Net. 
For fair comparisons, following previous work we use 10-fold cross-validation as the data split strategy and the precision (P), recall (R), and F1 score (F1) as evaluation metrics.
Meanwhile, we also verify the performance of two auxiliary tasks: emotion extraction (EE) and cause extraction (CE), using the same evaluation metrics as ECPE.

\subsection{Implementation Details}

We apply PyTorch to implement our framework.\footnote{https://pytorch.org}
We leverage pre-trained language model BERT \cite{devlin-etal-2019-bert} as our embedding layer.\footnote{The version of BERT is bert-base-chinese.}
We employ one-layer PFN \cite{yan2021partition} with hidden size of 300. 
Besides, the hyperparameters $\lambda_1$ and $\lambda_2$  are both set to 0.4. 
We set the batch size and the learning rate to 4 and 2e-5, respectively. 
We apply AdamW \cite{DBLP:journals/corr/abs-1711-05101} to optimize our model parameters. 
To prevent overfitting, the dropout rate is set to 0.1.

\subsection{Baselines}
In order to verify the effectiveness of our proposed model A$^2$Net, we compared it with the following strong methods. 
For a fair comparison, they all use BERT as the encoder.

\begin{itemize}
\setlength{\itemsep}{2pt}
\setlength{\parsep}{2pt}
\setlength{\parskip}{2pt}
    \item \textbf{ANTS} \cite{yuan-etal-2020-emotion}: ANTS solves ECPE with a sequence labeling approach and proposes a tagging scheme.
    \item \textbf{TransECPE} \cite{fan2020transition}: TransECPE is a transition-based framework that converts ECPE into a parsing-like directed graph construction task.
    \item \textbf{ECPE-2D} \cite{ding2020ecpe}: ECPE-2D leverages clauses pairs to construct a 2D representation matrix which integrated with auxiliary task predictions for ECPE task. 
    \item \textbf{PairGCN} \cite{chen2020end}: This method models the dependency relations among clause pairs with graph convolution networks.
    \item \textbf{RANKCP} \cite{wei2020effective}: RANKCP tackles the ECPE task from a ranking perspective and uses graph attention to model the inter-clause relations.
    \item \textbf{ECPE-MLL} \cite{ding2020end}: ECPE-MLL converts the ECPE task into the emotion-pivot cause extraction and the cause-pivot emotion extraction using the sliding window strategy.
    \item \textbf{MGSAG} \cite{bao2022multi}: MGSAG constructs a multi-granularity semantic aware graph to deal with ECPE task, and it is the current SoTA approach.
\end{itemize}

\section{Experimental Results}

\subsection{Main Results}

Table \ref{tab:main reuslt} shows the comparison results of our model (A$^2$Net) with the strong baselines on the emotion-cause pair extraction (ECPE) task and two auxiliary tasks: emotion extraction (EE) and cause extraction (CE). 

For the ECPE task, it is clear that our model A$^2$Net achieves 1.13\% and 1.82\% F1-score improvement over MGSAG (the current best method) and ECPE-MLL, respectively. 
Further analysis, we can find that the above advantage mainly comes from the improvement of the recall. 
Compared with MGSAG and ECPE-MLL, our recall is increased by 4.59\% and 5.45\% respectively, which indicates that consistent prediction on the three tasks allows the A$^2$Net model to detect more emotion-cause pairs under a considerable precision.

For auxiliary tasks, on the EE task, our model achieves 3.63\% and 1.94\% F1 improvement over MGSAG and ECPE-MLL, and 0.23\% F1 improvement over the previous best model RANKCP.
On the CE task, our model yields a great improvement of F1 scores by 1.23\% in comparison with the top-performing baseline MGSAG, and achieves 2.20\% and 2.05\% F1 improvement over RANKCP and ECPE-MLL.

We argue that all improvements come mainly from our proposed feature-task alignment module and inter-task alignment module. 
Both alignment mechanisms are able to collaboratively improve the performances of all tasks, and enhance the robustness of the model.
In the following part we performed corresponding experiments to verify our ideas.

\begin{table}[!t]
\centering
\begin{tabular}{lccc}
\toprule
& ECPE & EE & CE \\
\midrule
A$^2$Net (ours) & \textbf{76.34} & \textbf{90.80} & \textbf{78.35}  \\ \cdashline{1-4}
\ w/ Shared encoding & 69.97 & 84.81 & 72.66 \\
\ w/ Parallel encoding & 75.59  & 89.75 & 78.03 \\
\bottomrule
\end{tabular}

\caption{Performances (F1) with different feature encoding schemes.
} 
\label{tab:Effectiveness of Encoder}
\end{table}

\subsection{Effect of Feature-task Alignment}
The feature-task alignment module is capable of generating efficiency and independent task-specific and interactive features.
To verify the effectiveness of our feature-task alignment, we replaced the PFN with two encoding schemes: shared encoding and parallel encoding. 
The results are shown in Table \ref{tab:Effectiveness of Encoder}. 

In terms of the shared encoding, we encode the emotion features and cause features using a shared BiLSTM, in which emotion and cause features are entangled.
For parallel encoding, we utilize two BiLSTMs to capture emotion features and cause features separately, in which interaction information among different tasks is not considered. 
Firstly, we observe that the model with parallel encoding significantly outperforms the shared encoding among three tasks, indicating that it is important for the model to consider task-specific features. 
Furthermore, we can see that our model enjoys better performances when we consider both task-specific features and shared interaction features, compared with the parallel encoding.
This shows the necessity of aligning feature spaces for different tasks.

\subsection{Effect of Inter-task Alignment}

In this section, we investigate the effect of the inter-task alignment (ITA) mechanism and auxiliary tasks for A$^2$Net, and the results are plotted in Table \ref{tab:ablation reuslt}.

We first analyze the effect of the aligned direction of the inter-task alignment mechanism.
When we merely apply unidirectional alignment to regulate the predictions between ECPE and two auxiliary tasks, we can observe slight performance drops on three tasks to some extent.
Furthermore, after removing the inter-task alignment mechanism (bidirectional alignment), we find that the overall decreases in F1 score on three tasks happen, and are more than the any unidirectional alignment, which verifies the helpfulness of the alignment in label spaces among tasks, and bi-direction of alignment are more important for our model.

Besides, we also explore the effectiveness of auxiliary tasks, EE and CE.
It should be noted that inter-task alignment module does not work after the auxiliary tasks are removed.
When the auxiliary tasks are further removed, we can see that the model performances drop significantly, demonstrating that the auxiliary task can effectively contribute to the ECPE task.

\begin{table}[!t]
\centering

\begin{tabular}{lccc}
\toprule
& ECPE  &  EE &  CE \\
 \midrule
 A$^2$Net (ours) & \textbf{76.34} & \textbf{90.80}  & \textbf{78.35}  \\ 
 \cdashline{1-4}
 \ w/o ECPE$\rightarrow$EE$\times$CE & 75.83 & 90.65 & 78.05  \\
 \ w/o EE$\times$CE$\rightarrow$ECPE & 75.50 & 90.54 & 77.60 \\
 \ w/o ITA & 75.32 & 90.05 & 77.37  \\
\cdashline{1-4}
 \ w/o EE \& CE & 74.39 & - & - \\
 \bottomrule
\end{tabular}

\caption{Ablation study of inter-task alignment module and auxiliary task (F1). The ECPE$\rightarrow$EE$\times$CE means we use the prediction distribution of ECPE to align to EE$\times$CE (i.e., $\text{KL}(\tilde{y}_{ij}^{p}||\hat{y}^{p}_{ij})$ in Eq.\ref{KL}), and vice versa.} 
\label{tab:ablation reuslt}
\end{table}

\begin{table*}[!t]
\centering
\resizebox{1.0\textwidth}{!}{
\begin{tabular}{lccc}
\midrule
\textbf{Document} & \textbf{A$^2$Net(w/o ITA)} & \textbf{A$^2$Net} & \textbf{Ground-truth} \\
\midrule
\makecell[l]{
... The police visited the villagers of Nanyuan Village (c3), and they \\ 
learned that Meng was playing mahjong at a mahjong parlor opposite \\ 
his home the day before the incident (c4), through inquiries (c5), \textcolor{blue}{it wa-} \\
\textcolor{blue}{-s found that only Wang from the same village had gone out to an unk-} \\
\textcolor{blue}{nown destination}(c6), \textcolor{orange}{which aroused the} \textcolor{orange}{suspicion of the police} (c7).} & \makecell[c]{ECPE:\textcolor[RGB]{50,205,50}{[c7, c6]} \\ EE:\textcolor{red}{[]} \\ CE:\textcolor[RGB]{50,205,50}{[c6]}}  & \makecell[c]{ECPE:\textcolor[RGB]{50,205,50}{[c7, c6]} \\ EE:\textcolor[RGB]{50,205,50}{[c7]} \\ CE:\textcolor[RGB]{50,205,50}{[c6]}}  & \makecell[c]{ECPE:\textcolor[RGB]{50,205,50}{[c7, c6]} \\ EE:\textcolor[RGB]{50,205,50}{[c7]} \\ CE:\textcolor[RGB]{50,205,50}{[c6]}} \\
\midrule
\makecell[l]{On March 14 (c1), a magnitude 4.3 earthquake occurred in Yingquan \\
District, Fuyang City, Anhui (c2). Then (c3), \textcolor{blue}{a rumor of a magnitude} \\ 
\textcolor{blue}{6.8 earthquake occurred in Fuyang City at 2:15 am on March 15. (c4), }\\ 
\textcolor{orange}{which caused people to panic (c5)}...} & \makecell[c]{ECPE:\textcolor[RGB]{50,205,50}{[c5, c4]} \\ EE:\textcolor[RGB]{50,205,50}{[c5]} \\ CE:\textcolor{red}{[c2]}, \textcolor[RGB]{50,205,50}{[c4]}}  & \makecell[c]{ECPE:\textcolor[RGB]{50,205,50}{[c5, c4]} \\ EE:\textcolor[RGB]{50,205,50}{[c5]} \\ CE:\textcolor[RGB]{50,205,50}{[c4]}} & \makecell[c]{ECPE:\textcolor[RGB]{50,205,50}{[c5, c4]} \\ EE:\textcolor[RGB]{50,205,50}{[c5]} \\ CE:\textcolor[RGB]{50,205,50}{[c4]}}  \\
\midrule
\makecell[l]{Mr. Feng said frankly (c1), Jingjing is naughty on weekdays (c2), and \\
sometimes he is not polite (c3), \textcolor{blue}{but when it comes to the reason for th-}\\
\textcolor{blue}{-is injury}(c4), \textcolor{orange}{he can't hide his anger} (c5), just because of my son Dra- \\
-nk other children's yogurt (c6). Teacher Xing lost her mind (c7), she \\
was emotionally out of control (c8), then pulled the child out of the do-\\
-or (c9), the child was injured when the door was closed (c10)...} & \makecell[c]{ECPE:\textcolor[RGB]{50,205,50}{[c5,c4]},\textcolor{red}{[c5,c6]} \\ EE:\textcolor[RGB]{50,205,50}{[c5]} \\ CE:\textcolor[RGB]{50,205,50}{[c4]}, \textcolor{red}{[c6]}}  & \makecell[c]{ECPE:\textcolor[RGB]{50,205,50}{[c5,c4]} \\ EE:\textcolor[RGB]{50,205,50}{[c5]} \\ CE:\textcolor[RGB]{50,205,50}{[c4]}}  & \makecell[c]{ECPE:\textcolor[RGB]{50,205,50}{[c5,c4]} \\ EE:\textcolor[RGB]{50,205,50}{[c5]} \\ CE:\textcolor[RGB]{50,205,50}{[c4]}} \\
\midrule
\end{tabular}
}
\caption{Two examples for the case study.
The words in \textcolor{orange}{orange} are the emotion clause, and the words in \textcolor{blue}{blue} are the cause clause. The \textcolor[RGB]{50,205,50}{green} means correct predictions, \textcolor{red}{red} means wrong predictions.}
\label{tab:case study} 
\end{table*}

\subsection{Analysis of Prediction Consistency Cross Tasks}

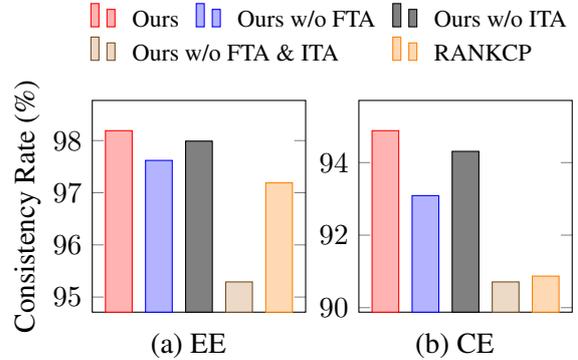
\begin{figure}[!t]
\centering
\begin{tikzpicture}
  \begin{axis}[name=plot1,ybar=5pt,height=0.57\columnwidth,width=0.57\columnwidth,ylabel=Consistency Rate (\%),xlabel=(a) EE,xtick style={draw=none},enlargelimits=0.2,xticklabel style = {yshift=5pt},ylabel shift=-3pt,xticklabels={,,},compat=newest,
  legend columns = 3,
  legend style = {at={(1.12,1.5)}, anchor=north, inner sep=3pt, style={column sep=0.15cm}, draw=none, font=\small},
  legend cell align=left,
  xlabel style={at={(0.45,-0.05)}}]
    \addplot[red,fill=red!30!white] coordinates {(0,98.19)};
    \addlegendentry{Ours}
    \addplot[blue,fill=blue!30!white] coordinates {(0,97.62)};
    \addlegendentry{Ours w/o FTA}
    \addplot[black,fill=gray] coordinates {(0,97.99)};
    \addlegendentry{Ours w/o ITA}
    \addplot[brown!60!black,fill=brown!30!white] coordinates {(0,95.29)};
    \addlegendentry{\makebox[0pt][l]{Ours w/o FTA \& ITA}}
    \addlegendimage{empty legend}
    \addlegendentry{}
    \addplot[orange,fill=orange!30!white] coordinates {(0,97.19)};
    \addlegendentry{\makebox[0pt][l]{RANKCP}}
  \end{axis}
  \begin{axis}[name=plot2,at={($(plot1.east)+(0.7cm,0)$)},anchor=west,ybar=5pt,height=0.57\columnwidth,width=0.57\columnwidth,xtick style={draw=none},enlargelimits=0.2,xticklabel style = {yshift=5pt},ylabel shift=-5pt,xticklabels={,,},compat=newest,xlabel=(b) CE,xlabel style={at={(0.45,-0.05)}}]
    \addplot[red,fill=red!30!white] coordinates {(0,94.88)};
    \addplot[blue,fill=blue!30!white] coordinates {(0,93.09)};
    \addplot[black,fill=gray] coordinates {(0,94.31)};
    \addplot[brown!60!black,fill=brown!30!white] coordinates {(0,90.71)};
    \addplot[orange,fill=orange!30!white] coordinates {(0,90.87)};
  \end{axis}
\end{tikzpicture}
\caption{Consistency of ECPE and EE (a), as well as CE (b).
}
\label{fig:consistency}
\end{figure}

In order to verify the effect of our proposed feature-task alignment module and inter-task alignment module on model prediction consistency among tasks, we conduct the experiments with multiple variants of A$^2$Net and the baseline RANKCP, as shown in Figure \ref{fig:consistency}. 
Specifically, we disassemble the emotion-cause pairs into a set of emotion clauses and a set of cause clauses which are considered as gold labels for EE and CE, respectively.
We calculated the consistency rate to evaluate the consistency of EE and CE tasks with ECPE task predictions by (EE \& ECPE) / ECPE or (CE \& ECPE) / ECPE, where EE, CE and ECPE denote the prediction results of corresponding tasks and \& denotes the logic AND.

On the EE task, we can find that when our model A$^2$Net removes the FTA and ITA modules, the consistency rate drops significantly, but when our model only removes FTA or ITA, the consistency rate decreases slightly. 
Furthermore, all of them outperform RANKCP, which indicates that FTA and ITA have well aligned EE with ECPE.

On the CE task, we can find that a dramatic drop occurs when we remove the FTA and ITA modules.
The consistency rate decreases slightly when only ITA is removed, while the consistency rate decreases more when only FTA is removed, indicating that our feature-level alignment is more effective for CE tasks. 
Moreover, all of them receive better consistency than RANKCP, indicating that both FTA and ITA are able to align CE and ECPE.

\subsection{Case Analysis}

Finally, to better understand the capacity of our proposed model, we empirically perform case study on EE, CE and ECPE tasks. 
Specifically, we demonstrate some predictions based on three instances randomly selected from testset, as shown in Table \ref{tab:case study}. 

In the first example, our A$^2$Net without ITA correctly predicts the emotion cause pair $(c_7, c_6)$ on the ECPE task and incorrectly on the EE and CE tasks. 
In contrast, our A$^2$Net model correctly predicts all ECPE, EE, and CE tasks after going through the inter-task alignment module.
In the second example, A$^2$Net without ITA correctly predicts the emotion cause pair $(c_5, c_4)$ on the ECPE task and incorrectly on the CE tasks.
However, A$^2$Net model correctly predicts all EE, CE and ECPE tasks.
In the third example, A$^2$Net without ITA predicts correctly in the EE task and incorrectly detects the cause clause $c_6$ in the CE task.
Meanwhile, the emotion-cause pair $(c_5, c_6)$ is predicted incorrectly. 
Nonetheless, after the inter-task alignment module, all predictions were correct in both the ECPE and CE tasks. 

This shows that after aligning between tasks, the model can identify emotion-corresponding causes, which is like the role of our proposed soft mask score.
We find that these cases are common in our dataset, which ultimately leads directly to an improvement in our model performance. 
This also demonstrates that our inter-task alignment module can normalize the inter-task training to make the model performs better and more stable.

\subsection{Model Efficiency Analysis}
In order to test the efficiency of our model, we conduct the experiments with multiple variants of A$^2$Net and the baseline RANKCP, and the results are plotted in
Table \ref{tab:Efficiency Analysis}.

In terms of parameter quantity, our A$^2$Net model is even less than RANKCP, and it can be found that since the ITA module does not introduce additional parameters, w/o ITA does not change the parameter quantity. 
In terms of inference speed, our model has the same inference speed as RANKCP, which shows that the efficiency of our model does not decrease due to the addition of the alignment mechanism.

\begin{table}[!t]
\centering
\begin{tabular}{lcc}
\toprule
Model & \#Param & Speed(doc/s) \\
\midrule
\ RANKCP & 105.97M & 195 \\
A$^2$Net (ours) & 104.97M & 195 \\ \cdashline{1-3}
\ w/o ITA & 104.97M  & 195 \\
\bottomrule
\end{tabular}

\caption{Parameter number and inference speed comparisons on ECPE. All models are tested with batch size 4.} 
\label{tab:Efficiency Analysis}
\end{table}

\section{Conclusion}

Existing best-performing ECPE works extensively leverage EE and CE as auxiliary tasks for better feature learning via multi-task learning (MTL).
In this paper, we further enhance the existing best-performing MTL-based ECPE by proposing feature-task alignment and inter-task alignment mechanisms.
At the feature space, the feature-task alignment mechanism aligns the task-specific features and the shared interactive feature with corresponding tasks.
At the label space, the inter-task alignment mechanism reduces the inconsistency among the predicted labels of EE, CE and ECPE. 
Experimental results on the benchmark ECPE data demonstrate the effectiveness of our methods. 
Further analysis shows that our system achieves better consistency than existing baselines, which explains the improvements of our model. 
The idea to align the feature space and label space in MTL framework is promising.
In the future work, we consider further constructing intra-clause relations, inter-clause relations, and relations among pairs of clauses for ECPE.

\section*{Acknowledgements}

This work is supported by the National Natural Science Foundation of China (No. 62176187), the National Key Research and Development Program of China (No. 2017YFC1200500), the Research Foundation of Ministry of Education of China (No. 18JZD015), the Youth Fund for Humanities and Social Science Research of Ministry of Education of China (No. 22YJCZH064), the General Project of Natural Science Foundation of Hubei Province (No.2021CFB385). 
This work is also the research result of the independent scientific research project of Wuhan University, supported by the Fundamental Research Funds for the Central Universities.

\bibliography{anthology}
\bibliographystyle{acl_natbib}

\end{CJK}
\end{document}